\documentclass[letterpaper, 10 pt, conference]{ieeeconf}  % Comment this line out if you need a4paper

\IEEEoverridecommandlockouts                              % This command is only needed if 
\usepackage{graphics} % for pdf, bitmapped graphics files
\usepackage{amsmath} % assumes amsmath package installed
\usepackage{cite}
\usepackage{subfigure}
\usepackage{tabulary}
\usepackage{multirow}
\usepackage{todonotes}
\graphicspath{{figs/}}
\usepackage[colorlinks,linkcolor=blue,anchorcolor=blue,citecolor=red]{hyperref} 
\usepackage{wrapfig,lipsum,booktabs}
\usepackage{xcolor}
\usepackage{color}
\usepackage{cite}
\usepackage{algorithmic} %format of the algorithm 
\usepackage[ruled,vlined,commentsnumbered]{algorithm2e}

\usepackage[flushleft]{threeparttable}
\usepackage{gensymb}  % degree symbol
\usepackage{amsmath}
\usepackage{amssymb}

\usepackage{flushend}

\graphicspath{{figs/}}

\usepackage{siunitx}
\usepackage{url}   % url link

\title{\LARGE \bf
Obstacle Avoidance through Deep Networks based Intermediate Perception
}

\author{Shichao Yang*, Sandeep Konam*, Chen Ma, Stephanie Rosenthal, Manuela Veloso, Sebastian Scherer % <-this % stops a space
\thanks{The Robotics Institute, Carnegie Mellon University, 5000 Forbes Ave, Pittsburgh, PA 15213, USA.
		{\tt\small \{shichaoy, skonam\}@andrew.cmu.edu}}
\thanks{* The first two authors contributed equally.}
}

\begin{document}

\maketitle
\thispagestyle{empty}
\pagestyle{empty}

%%%%%%%%%%%%%%%%%%%%%%%%%%%%%%%%%%%%%%%%%%%%%%%%%%%%%%%%%%%%%%%%%%%%%%%%%%%%%%%%
\begin{abstract}
	%%%%%%%%% ABSTRACT
% importance of research
% Problem descripttion, scope of research
% Existing Approach, our approach
% Results
Obstacle avoidance from monocular images is a challenging problem for robots. Though multi-view structure-from-motion could build 3D maps, it is not robust in texture-less environments. Some learning based methods exploit human demonstration to predict a steering command directly from a single image. However, this method is usually biased towards certain tasks or demonstration scenarios and also biased by human understanding. In this paper, we propose a new method to predict a trajectory from images. We train our system on more diverse \textit{NYUv2} dataset. The ground truth trajectory is computed from the designed cost functions automatically. The Convolutional Neural Network perception is divided into two stages: first, predict depth map and surface normal from RGB images, which are two important geometric properties related to 3D obstacle representation. Second, predict the trajectory from the depth and normal. Results show that our intermediate perception increases the accuracy by $20\%$ than the direct prediction. Our model generalizes well to other public indoor datasets and is also demonstrated for robot flights in simulation and experiments.

%From simulation and experiments, robots can navigate safely in various challenging environments using our approach.

%More than $95\%$ of the predicted trajectory doesn't hit obstacles on \textit{NYUv2} dataset.

%Also, the predicted trajectory is suitable for precise robot control in a short time.
%Besides, it only generates current steering command instead of trajectory so it requires the prediction to be fast.

	\label{Abstract}
\end{abstract}

\section{Introduction}
\label{sec:Introduction}
Autonomous vehicles have raised wide interest in recent years with various applications such as inspection, monitoring and mapping. To navigate autonomously, the vehicles need to detect and avoid 3D obstacles in real time. Various range sensors such as laser\cite{bachrach2009autonomous}, stereo cameras\cite{shen2013vision}, and RGB-D depth cameras\cite{fang2016robust} could be used to build 3D map of the environment. In this work, we focus on more challenging monocular obstacle avoidance. A typical approach to vision-based navigation problem is to use geometric 3D reconstruction techniques such as structure from motion (SfM) and simultaneous localization and mapping (SLAM). They usually track visual features to build sparse \cite{mur2015orb} or semi-dense 3D map \cite{engel2014lsd}. However, monocular vSLAM may not be robust especially in challenging low-texture or degenerated environments. Besides, their map is usually relatively sparse and cannot fully represent 3D obstacles.

%but these sensors are usually heavy and expensive compared to a single monocular camera especially for weight constrained micro aerial vehicles (MAVs). 

%There have been many impressive research on the obstacle avoidance and autonomous navigation of ground and aerial vehicles. Traditional systems usually utilize range-sensors such as laser \cite{bachrach2009autonomous}, stereo cameras \cite{shen2013vision} and RGB-D \cite{fang2016robust} cameras to build a 3D map using SfM and SLAM \cite{mur2015orb} \cite{engel2014lsd} then uses sampling based planner \cite{karaman2011sampling} or optimization based planner \cite{zucker2013chomp} to generate trajectories. For miniature MAVs \cite{brockers2014towards}, due to the payload constraints, monocular cameras are usually the only choice. However, monocular SLAM is not robust in some challenging indoor environments such as long corridors with homogeneous surface due to a lack of sufficient feature points to track. It usually outputs a sparse or semi-dense 3D map which doesn't fully represent the full 3D obstacle surfaces and conveys little information for motion planning.

\begin{figure}[t]
  \centering
   \includegraphics[scale=0.27]{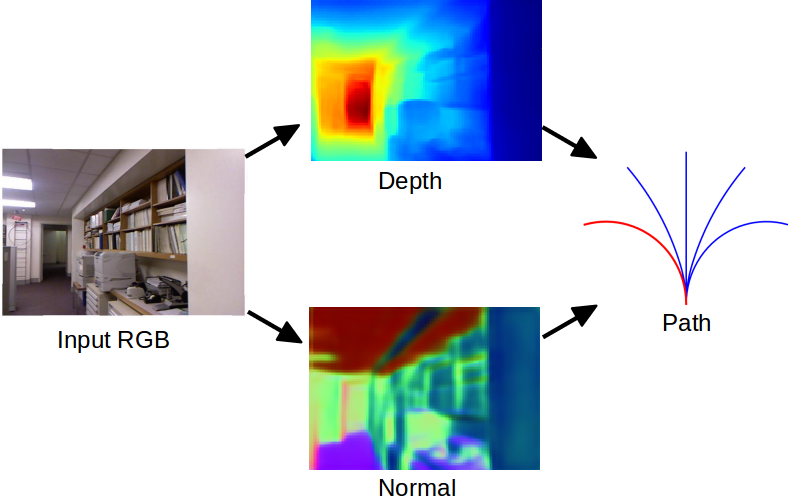}
   \caption{Method Overview. Instead of directly predicting path from RGB image, we propose intermediate perception:
first predict depth and surface normal which are closely related to 3D obstacles, then predict the path from the depth and normal. Both steps utilize CNN.}
   \label{fig:method overview}
\end{figure}

On the other hand, for the image in Fig. \ref{fig:method overview}, a human can easily understand the 3D layout with obstacles and generate the best motion commands. Recently, to mimic human's behavior, there are also some learning based algorithms to predict paths directly from raw RGB image especially with the popularity of Convolutional neural networks (CNNs) \cite{chen2015deepdriving} \cite{giusti} \cite{taideep}. To train CNN models, a large dataset with motion command labels is needed. Previous work mainly utilized human demonstration to collect datasets with labels, but it has some problems such as being time-consuming to collect large demonstration data and usually biased to certain scenarios where demonstrations happen. For example, in Fig. \ref{fig:method overview}, people might select 'straight forward' or 'left turning' based on their own analysis of risk and task requirements. They only predict the instantaneous steering command (left, right, etc.) so it requires the learning algorithm to predict frequently. It is also difficult to evaluate whether robots hitting obstacles or not as it depends on the how long the steering command last.
%give steering commands very frequently otherwise it might hit obstacles using the old angular velocity. 

%One common problem of the above methods is that they all use human demonstration to collect ground truth datasets which have some drawbacks. Firstly, it is difficult to get a large-scale and diverse collection of training data. Secondly, their data usually is biased to certain scenarios where demonstrations happen and also biased by human understanding. 

%instead of the commonly used steering command (angular velocity)
In this paper, we propose to predict trajectories directly from one single image. Trajectories contain more 3D information than instant steering command. To do that, we first propose a new method to get ground truth labels from RGB-D images automatically without human demonstrations. Then we propose a two-stage CNN with intermediate perception. The first stage is to predict the depth and surface normal from images, which are two important geometric properties related to 3D obstacle representation. The second step is to predict a path from the depth and normal maps using another CNN model. In the latter experiments, we demonstrate that our method is more robust compared to direct CNN prediction from raw RGB images. An overview of the proposed perception method is presented in Fig. \ref{fig:method overview}.

In all, our main contributions are:
\begin{itemize}
\item
Generate ground truth trajectory label from images automatically based on the designed 3D cost functions instead of human demonstration limited to certain scenarios. 
%Our CNN model generalizes well to various environments.
%We use Euclidean cost map to compute the ground truth trajectory instead of collecting path from human demonstration which could greatly simplify the process.

\item
Propose novel idea of mediated perception to instruct CNN to learn depth and surface normal before path prediction. It is more accurate than direct prediction demonstrated in the experiments.
%The scene geometry could improve robot's understanding of the environments.
%
\item
Apply to robot simulation and real flight. Robot is able to avoid various configuration of obstacles.
%using the trained model on \textit{NYUv2}.

\end{itemize}

We first provide the related work in Section \ref{sec:Related}. Details of the proposed method including the dataset creation and CNN architecture is presented in Section \ref{sec:Method}. In Section \ref{sec:Experiments}, we present experiments on several public datasets, dynamic simulations and real flights. Finally, the paper is concluded in Section \ref{sec:Conclusion}.

\section{Related Work}
\label{sec:Related}
% multiple images, slam.
% learning based method.
There are two basic categories of methods that utilize monocular images to detect obstacles and generate motion commands. The first category is to analyse the object position and size based on monocular motion cues and low-level features. Mori \textit{et al.} detect SURF features and analyze their size change in consecutive frames to detect front obstacles without geometric assumption or training on the scenes but may not work for lateral and far away obstacles. Optical flow \cite{souhila2007optical}, vanishing points direction \cite{bills2011autonomous} and wall-floor intersection detection \cite{ok2012vistas} can also be utilized but they are usually suitable for relative easy tasks such as corridor following.

The second category is to directly predict 3D obstacle position or motion commands using learning algorithms especially recent CNNs. 3D scene layout understanding \cite{hedau2009recovering} \cite{chao2016pop} could be an efficient way of representing the environment but it is usually limited to Manhattan environments or corridor with clear boundaries \cite{syang2016popslam} which might not be satisfied in all environments. CNN is widely used in semantic segmentation \cite{long2014fully}, depth estimation \cite{eigen2015predicting}, surface normal prediction \cite{wang2015designing} and so on. Chen \textit{et al.} \cite{chen2015deepdriving} propose a 'deep driving' system to predict some features such as position on road and distance to car, then generate steering commands (angular velocity) using a feedback controller. Tai \textit{et al} \cite{taideep} apply CNN to ground robot navigation by directly predicting five steering direction from the input of depth image. Giusti \cite{giusti} \textit{et al} apply CNN to autonomous trail following using aerial vehicles. In \cite{mancini2016fast}, obstacles are directly detected and represented through deep network based depth prediction but it doesn't further apply to robot navigation. Recently, deep reinforcement learning (RL) with Q-learning also shows promising results. \cite{levine2016end} use a deep RL approach to directly map raw images to robot motor torques. However, most of RL work is demonstrated in simple or simulated environments such as video games and needs more time for real robotic applications.

%such as distance to surrounding vehicles and lane boundary, then

%to build another CNN model as the second stage to 
%We first predict depth and surface normal from raw RGB images, which could represent the rough 3D model.
%to generate ground truth trajectory using RGB-D images 
%by predicting command from raw RGB image
%but they do not have surface normal and they do not address trajectory generation while we do.
%Recently, there have been many works to understand 3D scenes through depth \cite{eigen}, surface normal \cite{wang2015designing} and layout understanding \cite{hedau2009recovering}, which can all be utilized to help detect 3D obstacles. Layout understanding usually requires manhattan assumption or wall-ground assumption \cite{syang2016popslam} which might not hold in all images. Depth estimation is the more straight forward and also general for various environments.

\section{Visual Perception}
\label{sec:Method}
We first propose an automatic method of labelling trajectory from image to create a large training dataset. Then CNN is utilized to predict depth and surface normal to form a rough 3D model. This model might be noisy, inaccurate, and even invalid in some cases so we can not use the standard motion planner \cite{karaman2011sampling} \cite{zucker2013chomp} to generate trajectories. Instead, we utilize the predicted 3D model to predict the trajectory in another CNN model. %Results show that this two-stage model performs better than directly predicting from RGB images.

\subsection{Create Dataset}
\label{sec:create dataset}
\begin{figure}[t]
\vspace{1.5 mm}
\begin{center}
\scalebox{0.38}{\includegraphics{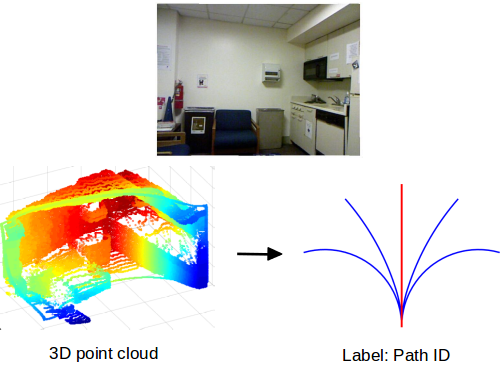}}
\end{center}
   \caption{Generating ground truth path label. Using the provided depth image, a 3D point cloud map is built. Then the best path in red is selected based on the Euclidean obstacle cost and path smoothness cost. }
\label{fig:depth to path}
\end{figure}

We use the \textit{NYUv2} RGB-D dataset which contains many diverse indoor scenes. For trajectory prediction, in theory, robots could execute numerous continuous 3D trajectories which will be a difficult high-dimensional regression problem. To simplify the problem, we change it into a classification problem by discretizing the continuous trajectory space into five categories of trajectories: left turn, left forward, straight forward, right forward, right turn, shown in the right of Fig. \ref{fig:method overview}. Then for each image, the best one is selected as the ground truth label. These five trajectories are generated by the maximum dispersion theory of Green \textit{et al} \cite{green2007toward}. Each path is around $2.5$m long. It is difficult for people to tell whether a trajectory hits obstacles or not from 2D images so we design a cost function to automatically select the best path. We first project the provided depth image into 3D point cloud then build a 3D Euclidean distance map. The cost function is commonly used in optimization based motion planner \cite{zucker2013chomp} \cite{fang2016robust}. An example of the 3D point cloud and selected label is shown in Fig. \ref{fig:depth to path}. In more detail, the cost function of a trajectory $\xi$ is computed by:
%We could manually label the best path, however, the label might be affected by different people . To avoid the ambiguity, 
%avoid the ambiguity of manual labelling as explained in Section \ref{sec:Related} and design a cost function to automatically select the best path. 

\begin{equation}
\label{eq:problemstatement}
   J(\xi) = f_{obst}(\xi) + w f_{smooth}(\xi)
\end{equation}
\noindent where $w$ is the weighting parameter. $f_{obst}(\xi)$ is the obstacle cost function\cite{zucker2013chomp} penalizing closeness to obstacles:

\begin{equation}
f_{obst}(\xi) = \int_0^1 c_{obs}(\xi(t)) \| \frac{d}{dt} \xi(t) \| dt
\end{equation}
\noindent where $c_{obs}(\xi(t))=\|\max(0,d_{max}-d(\xi(t))\|^2$. $d_{max}$ is the maximum distance upon which obstacle cost is available. For $NYU$ dataset, we set $d_{max}=3.5$m. $d(\xi(t))$ is the distance to obstacles from the distance map. $f_{smooth}(\xi)$ measures the smoothness of the path and penalizes the high derivatives:

\begin{equation}
f_{smooth}(\xi)=\frac{1}{2} \int_0^1 \| \frac{d}{dt} \xi(t) \|^2 dt
\end{equation}

The final ground truth label distribution for \textit{NYUv2} is shown in Table \ref{table: nyu label distri}. We can see that five categories are nearly equally distributed and thus fair for the latter evaluation and comparison of different methods.

\begin{table}
\vspace{1.5 mm}
\caption{Trajectory label distribution on \textit{NYUv2} dataset.}
\begin{center}
  \begin{tabular}{c c c c c c}
	\toprule
	Class ID      		&Distribution \\ \midrule
	Left turn    		&20.70\% \\
	Left forward  		&17.08\% \\	
	Straight forward  	&22.15\% \\	
	Right forward 		&18.01\% \\ 	
	Right turn  	     	&22.08\% \\
	\bottomrule
  \end{tabular}
\end{center}
\label{table: nyu label distri}
\end{table}

\subsection{Intermediate Perception - depth and surface normal}
\label{sec:pred depth}
Different from existing CNN based navigation methods \cite{giusti} \cite{taideep} to predict the command directly from RGB image, we first predict depth and surface normal. As explained in Section \ref{sec:Related}, there are many 3D geometry understanding methods that could help obstacle avoidance and we choose depth estimation due to its generality for various environments. Surface normals are also useful for navigation for example in Fig. \ref{fig:method overview}, the predicted normal on the right half of the image point to the left, it will give some information to the robot that there is a right wall and it is better to turn left.
%which are two important geometry properties related to 3D obstacle avoidance

There has been lots of work in depth and normal estimation from a single image \cite{eigen2015predicting} \cite{wang2015designing}. We utilize the CNN model from Eigen \cite{eigen2015predicting}, which is a multi-scale fully convolutional network. It first predicts a coarse global output based on the entire image area, then refines it using finer-scale local networks. The cost functions for depth training are defined as follows: suppose $d$ is the log difference between predicted and ground depth, then depth loss is:
$$L_{depth}=\frac{1}{n} \sum_i d_i^2 - \frac{1}{2n^2} \left( \sum_i d_i \right)^2 + \frac{1}{n} \sum_i \triangledown d_i^2 $$
where $\triangledown d_i$ is the gradient magnitude of $d$. If the ground truth and predicted normal vector are $v$ and $v^*$, then the dot product between vectors is used to define the normal loss function:
$$L_{normal}=- \frac{1}{n} \sum_i v_i \cdot v_i^* $$

\subsection{Trajectory prediction}
\label{sec:pred path}
\begin{figure*}
\begin{center}
  \scalebox{.65}{\includegraphics{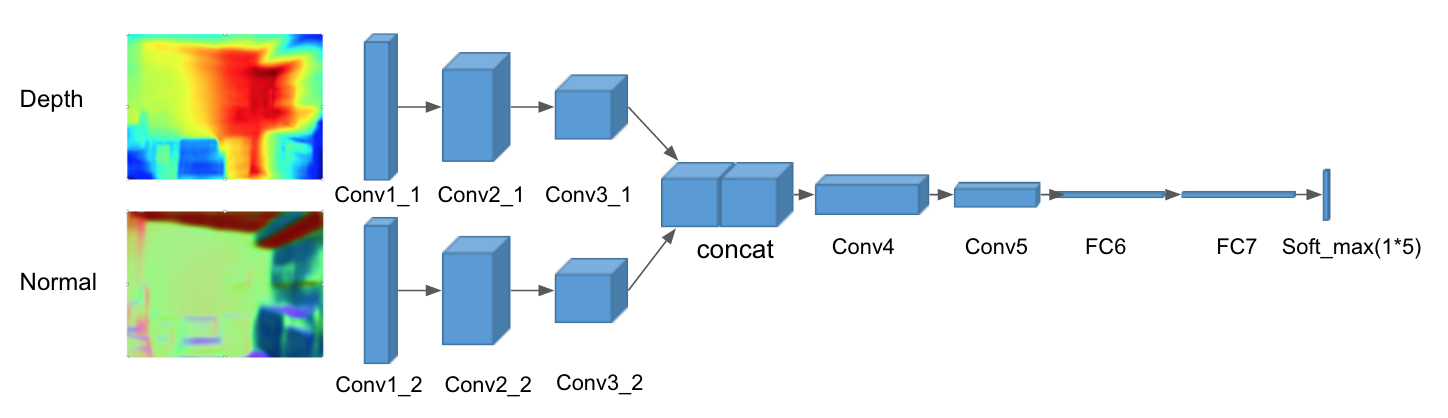}}
\end{center}
\caption{Proposed model architecture to predict path from depth and surface normal. It has two branches at the beginning to receive two input information. The prediction result is a label within five classes. For predicting depth and normal images, we use the model of \cite{eigen2015predicting}.}
\label{fig:path cnn}
\end{figure*}

We design another CNN network in Fig. \ref{fig:path cnn} to utilize the predicted depth and normal to get the final path classification. It is a modification of standard Alexnet \cite{krizhevsky2012imagenet} with two inputs. For symmetry, we replicate the depth image (one channel) into three channels. More complicated HHA feature \cite{gupta2014learning} from depth image could also be used. Then depth and normal images learn convolution filters separately and are fused at the fourth layer. By doing this, the final prediction will merge two sources of information. For training, we minimize the standard classification cross-entropy loss:
%The predicted depth and normal image might be inaccurate, noisy and even invalid (eg. negative depth value), we cannot directly project them into 3D space to select the best path. Instead,
\begin{small}
\begin{equation}
\label{cnn_loss}
L(C,C^*)=-\frac{1}{n} \sum_i C^*_i \log(C_i)
\end{equation}
\end{small}

\noindent where $C_i=e^{z_i}/\sum_c e^{z_{i,c}}$ is the softmax class probability given the CNN convolution output $z$.

%Similar as FlowNet in ~\cite{flownet},

\section{Experiments}
\label{sec:Experiments}
\subsection{Training and testing}
In the training phase, we use the created dataset in Section \ref{sec:create dataset} to train our two-stage CNN separately. In the first stage, the ground truth surface normal is computed by fitting local least-squares planes in the point cloud based on \cite{silberman2012indoor}. Actually, we can also directly use the public available CNN model of \cite{eigen2015predicting}. In the second stage, we use the \textit{ground truth} depth and surface normal as inputs to train CNN for path classification. We also augment the training data by flipping images horizontally. Note that for surface normal flipping, we need to reverse the horizontal normal component.

In the testing phase, we only use the raw RGB image as inputs of our two-stage CNN.

\subsection{NYUv2 dataset Evaluation}
The baseline for comparison is to directly train Alexnet CNN model to predict path label from raw RGB images, which has been adopted in many other CNN navigation works \cite{chen2015deepdriving} \cite{giusti}\cite{taideep}. We use the same settings of parameters for CNN weight initialization and training for comparison.

\subsubsection{Qualitative Results}
\begin{figure}[t]
\vspace{1.5 mm}
\begin{center}
\scalebox{.35}{\includegraphics{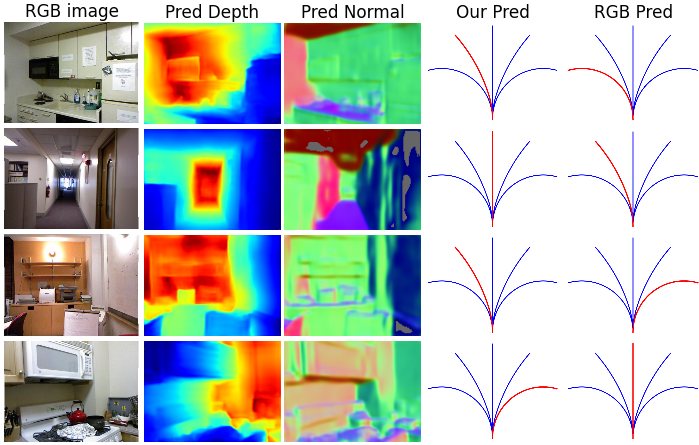}}
   %\includegraphics[width=0.8\linewidth]{egfigure.eps}
\end{center}
   \caption{Example of path prediction on \textit{NYUv2} dataset. The input is only RGB image. Our intermediate prediction and other direct prediction are shown in the last two columns in red color.  From left to right: RGB image, predicted depth image, predicted surface normal, and predicted paths. In the top image, two predictions are similar. In the bottom images, our method performs better.}
\label{fig:nyu path pred}
\end{figure}

A qualitative comparison between our method and baseline method is shown in Fig. \ref{fig:nyu path pred}. We can see that our method generates safer and more reasonable trajectory most of the time.

\subsubsection{Quantitative Results}
The result of the two methods is presented in Table \ref{table:nyu test error}. The '\textit{Accuracy}' column represents the standard classification accuracy. We also note that there is no clear distinction between different trajectories, for example, in some open space environments, going forward and turning both works.  Therefore, to make the evaluation more meaningful, we also check whether the predicted path lies in the top two best ground truth paths shown as '\textit{Top-2 accuracy}' column. From the table, our method with intermediate perception performs much better compared to direct perception from RGB image, with an accuracy increase of 20\%. It is also more accurate (42\% improvement) compared to random prediction.
%For example, both 'left turning' and 'left forward' is reasonable and has small difference for the scene in Fig .\ref{fig:method overview}.

\iffalse
\begin{table}
\caption{Comparison of path prediction on \textit{NYUv2} dataset.}
\begin{center}
\begin{tabular}{c c c c}
\toprule
Method       		& Accuracy   & Top-2 accuracy  & Safe prediction   \\  \midrule
Two-stage (Ours)    & 64.07\%    & 82.11\%	      & 92.08\% 		   \\  \midrule
Two-stage			&\multirow{2}{*}{60.34\%}  & \multirow{2}{*}{78.09\%}        &\multirow{2}{*}{90.68\%} \\ 
only depth (Ours) 												   \\  \midrule
CNN on RGB          & 39.20\%   & 60.19\%         & 78.73\%    	   \\  \midrule
Random			    & 20.00\%   & 40.00\%   	     & 62.38\%     		\\
\bottomrule
\end{tabular}
\end{center}
\label{table:nyu test error}
\end{table}
\fi

\begin{table}
\caption{Comparison of path prediction accuracy on \textit{NYUv2} dataset (\%).}
\begin{center}
\begin{tabular}{c c c c}
\toprule
Method       		& Accuracy   	 & Top-2 accuracy  & Safe prediction   \\ 
Ours				    & \textbf{64.07}	 & \textbf{82.11}  & \textbf{92.08} 		      \\ 
Our depth only		& 60.34	 	 	 & 78.09           & 90.68		      \\ 
Direct predict      & 39.20       	 & 60.19           & 78.73    	  \\
Random predict		& 20.23      	 & 40.00   	       & 62.38     		\\
\bottomrule
\end{tabular}
\end{center}
\label{table:nyu test error}
\end{table}

We also report the percentage of predicted paths hitting the 3D obstacles shown in the '\textit{Safe prediction}' column, which is an important metric related to robot's safety. This metric is not evaluated in other works because they usually predict instant steering command which cannot tell safety or not precisely. 92.08\% of the predicted paths using our proposed methods are safe. We also need to know that during actual robot applications, robots will continuously re-predict trajectories from new coming images instead of following one trajectory till ends. So we can consider obstacles only within a certain distance (eg. 2$m$) to the robots, the safe prediction ratio increases to 95.55\%. Note that since images are usually taken at some distance from obstacles, random prediction will not necessarily generate unsafe trajectory hitting obstacles all the time. 

%If we choose the one out of five paths randomly, 62.38\% of path are safe.

Results of our staged CNN model with only depth prediction can be seen in Table \ref{table:nyu test error}. It performs slightly worse compared to combining depth and surface normal. The confusion matrix of our method is shown in Table \ref{table:nyu confusion mat}.  We can see that most of the times, the prediction is correct and only a few times, the prediction will generate totally opposite direction. The reason why our network performs better is that we add ground truth depth information in the first stage's CNN model training which will guide the network to learn 3D scene information related to obstacle avoidance. However, the baseline method might not be able to learn this 3D information just from a single image without ground truth depth instruction.
%We also note a phenomenon that during the training process, the baseline method converges quickly in a few epochs leading to great over-fitting

\if(0)
\begin{table}
\caption{Comparison of testing accuracy on \textit{NYUv2} data.}
\begin{center}
\begin{tabular}{c c c c}
\toprule
Method       		&Accuracy  & Top-2 accuracy  &Safe prediction \\ \midrule
Intermediate (Ours) & 64.07\%  & 82.11\%	         &92.08\% \\
Intermediate, only depth (Ours) & 64.07\%  & 82.11\%	         &92.08\% \\
CNN on RGB          & 39.20\%  & 60.19\%          &78.73\% \\
Random			    & 20.00\%  & 40.00\%   	     &62.38\% \\
\bottomrule
\end{tabular}
\end{center}
\label{table:nyu test error}
\end{table}
\fi

\begin{table}
\caption{Confusion matrix for paths prediction on \textit{NYUv2}.}
\begin{center}
\begin{tabular}{c c c c c c}
\toprule
Class	   	& Left   & Left$+$ & Straight  & Right$+$  & Right \\ \midrule
Left			& 0.65   & 0.11    & 0.11 	  & 0.03 	  & 0.10  \\ 
Left$+$     & 0.14   & 0.64    & 0.12      & 0.04	  & 0.06  \\  
Straight    & 0.07   & 0.14    & 0.60      & 0.10      & 0.08  \\ 
Right$+$    & 0.07   & 0.05    & 0.21      & 0.56      & 0.11  \\ 
Right       & 0.09   & 0.02    & 0.06      & 0.10      & 0.73  \\ 
\bottomrule
\end{tabular}
\end{center}
\label{table:nyu confusion mat}
\end{table}

\subsection{Other Public Indoor dataset Evaluation}
\begin{figure}[t]
\vspace{1.5 mm}
\begin{center}
\scalebox{.4}{\includegraphics{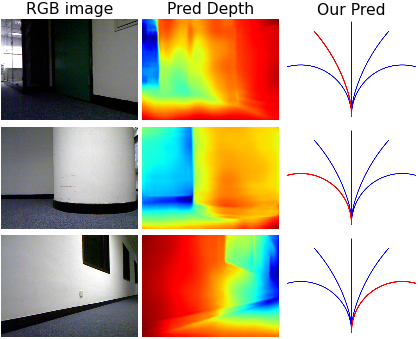}}
\end{center}
   \caption{Some prediction examples in \textit{Ram-lab} dataset images.}
\label{fig:hk path predict}
\end{figure}

Recently, Tai \textit{et al}\cite{taideep} propose the \textit{Ram-lab} RGB-D corridor dataset for CNN based autonomous navigation. We cannot directly compare with their method as they predict steering command from true depth image instead of RGB image. Besides, their dataset labels come from human demonstration, which is different from ours and also not even distributed. So we generate a new path label for their dataset using the same method and parameters in Section \ref{sec:create dataset}. Depth image is pre-processed by the cross-bilateral filter of Paris \cite{paris2006fast} to fill the missing regions in order to project to 3D space. Note that we directly apply the trained model from \textit{NYUv2} on this dataset without tuning any parameters. Some prediction examples are shown in Fig. \ref{fig:hk path predict}. A metric comparison on all the images is presented in Table \ref{table:hk test error}.
%and compute surface normals

We can find that our method still outperforms the baseline methods in terms of classification \textit{accuracy} and \textit{Safe prediction} ratio. However, the improvement is not as large as the \textit{NYUv2} dataset mainly because \textit{Ram-lab} dataset contains only corridor images with similar scenes structures and appearance.

\begin{table}
\caption{Comparison of path prediction on \textit{Ram-lab} datasets (\%).}
\begin{center}
\begin{tabular}{c c c c}
\toprule
Method 		        &Accuracy   &Top-2 accuracy   &Safe prediction \\ \midrule
Ours			        & \textbf{50.05}	   &\textbf{73.21}	          &\textbf{91.10} \\
Direct predict      & 42.68	   &65.67	          &87.92 \\
Random predict	    & 20.00	   &40.00  		     &72.37 \\
\bottomrule
\end{tabular}
\end{center}
\label{table:hk test error}
\end{table}

\subsection{Quadrotor simulation flight}
\begin{figure}[t]
\vspace{1.5 mm}
\begin{center}
\scalebox{.4}{\includegraphics{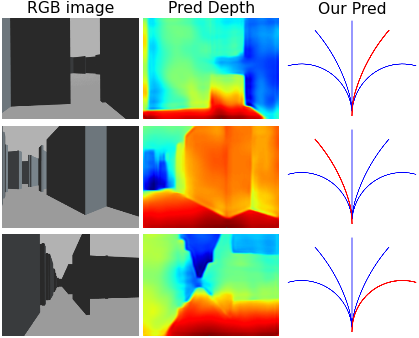}}
   %\includegraphics[width=0.8\linewidth]{egfigure.eps}
\end{center}
   \caption{Some prediction examples in gazebo simulations. }
\label{fig:gazebo prediction}
\end{figure}

\begin{figure}
\vspace{1.5 mm}
\begin{center}
\scalebox{.23}{\includegraphics{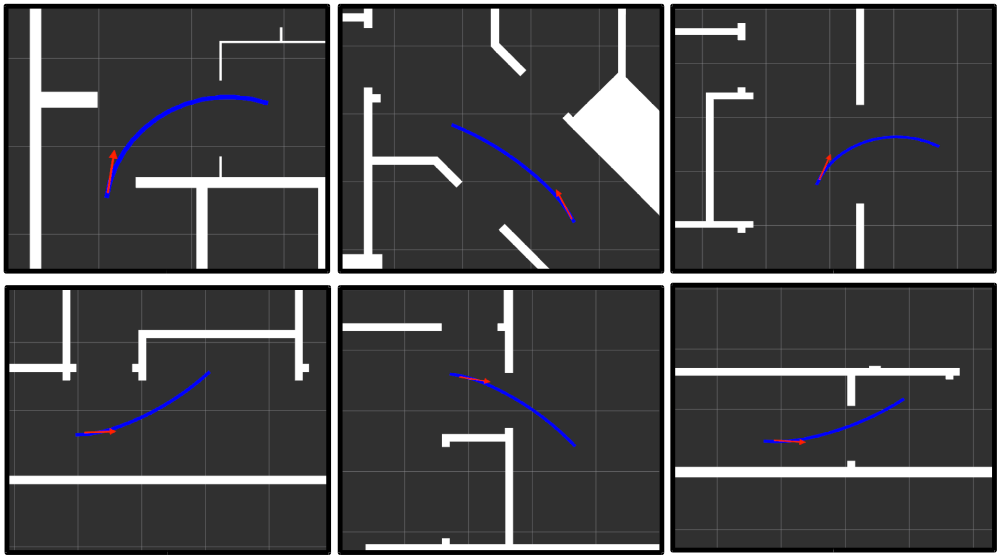}}
   %\includegraphics[width=0.8\linewidth]{egfigure.eps}
\end{center}
   \caption{Top view of path prediction in gazebo simulations. The red arrow represents the robot pose, the blue line is the predicted
   trajectory. The white areas in the image represents the obstacles.}
\label{fig:gazebo topview prediction}
\end{figure}

\begin{figure*}
\vspace{1.5 mm}
\begin{center}
  \scalebox{.4}{\includegraphics{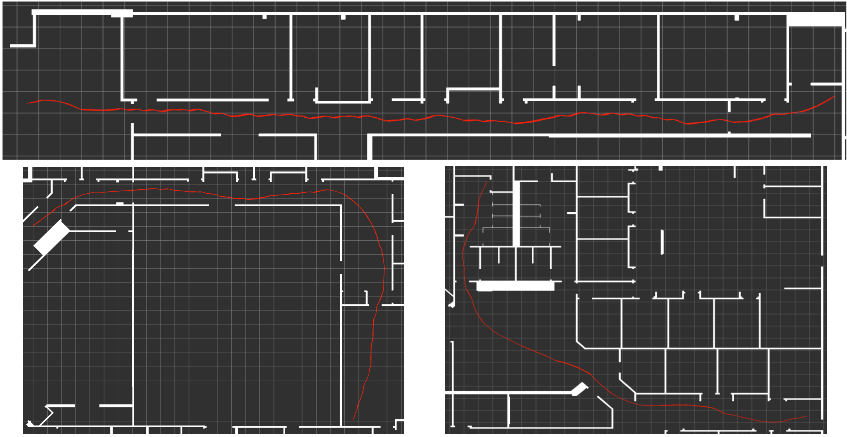}}
\end{center}
\caption{Some scenarios of the dynamic simulation. White area represents obstacles and red curve is the quadrotor history pose.}
\label{fig:simulation map}
\end{figure*}

We also evaluate our perception algorithm for continuous robot flight in simulations shown in Fig. \ref{fig:gazebo prediction}. Our CNN model predicts a trajectory in real-time, then the robot will follow the trajectory until a new predicted trajectory comes. Yaw heading of quadrotor is set as the tangent direction of the trajectory. We use the ardrone quadrotor dynamic simulator developed by Engel \textit{et al} \cite{engel2014scale} in the default gazebo willowgarage world model without texturing the surface. A simple PID controller is designed for trajectory tracking. As this is a whole navigation system, the final flight performance is affected by both the path prediction error and also the uncertainty in dynamic modelling and trajectory tracking. In the simulation, there are various kinds of door and wall configurations to test the robustness of our system. Some of the prediction examples are shown in Fig. \ref{fig:gazebo prediction} and the top-view trajectory prediction is shown in Fig. \ref{fig:gazebo topview prediction}. The whole history is shown in Fig .\ref{fig:simulation map}. We can see that even the scene is quite different from \textit{NYUv2} dataset, our CNN model still works well and the quadrotor is able to travel long distance and cross very narrow doors.

\vspace{1.5 mm}

\vspace{1.5 mm}

We also provide quantitative analysis of the simulation flight performance shown in Table \ref{table:analysis of simulation}. The ground truth pose is provided by gazebo. The mean robot to obstacle distance is 0.98m. The narrowest door is only 0.78m width while quadrotor width is 0.52m. The average prediction time of our whole system is $38.5$ms on a GeForce GTX 980 Ti GPU so it is able to run in real time over $25$Hz.

We need to note that our CNN model can only be used for local obstacle avoidance so the quadrotor can easily come to a dead end if there is no high level planning shown at the top scene of Fig. \ref{fig:simulation map}. Moreover, we currently only use five primitive paths which might not be enough for real world requirements. For example, if there is sharp turning, U-turn trajectory or stop primitives to select, robot flight could perform better. These techniques have been used in some actual flight system \cite{fang2016robust}.
%or the path should be shorter...

\vspace{1.5 mm}

\begin{table}
\caption{Statistic analysis of quadrotor simulation.}
\begin{center}
\begin{tabular}{c c c c}
\toprule
Method 		          & Scene I      \\ \midrule
Traveling distance    & 264.8 m 		  \\
Distance to obstacles & 0.98 m   	  \\
Prediction time       & 38.5 ms       \\
\bottomrule
\end{tabular}
\end{center}
\label{table:analysis of simulation}
\end{table}

\vspace{1.5 mm}

\subsection{Real quadrotor flight}
\begin{figure*}
\centering
\subfigure[]{
  \includegraphics[width=0.80\textwidth]{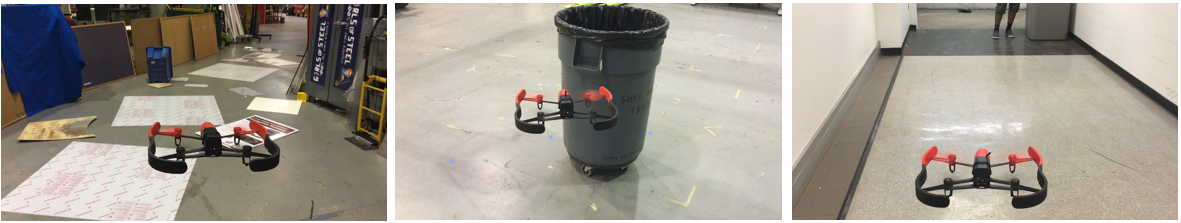}
  \label{fig:real scenes}
}
\subfigure[]{
  \includegraphics[width=0.80\textwidth]{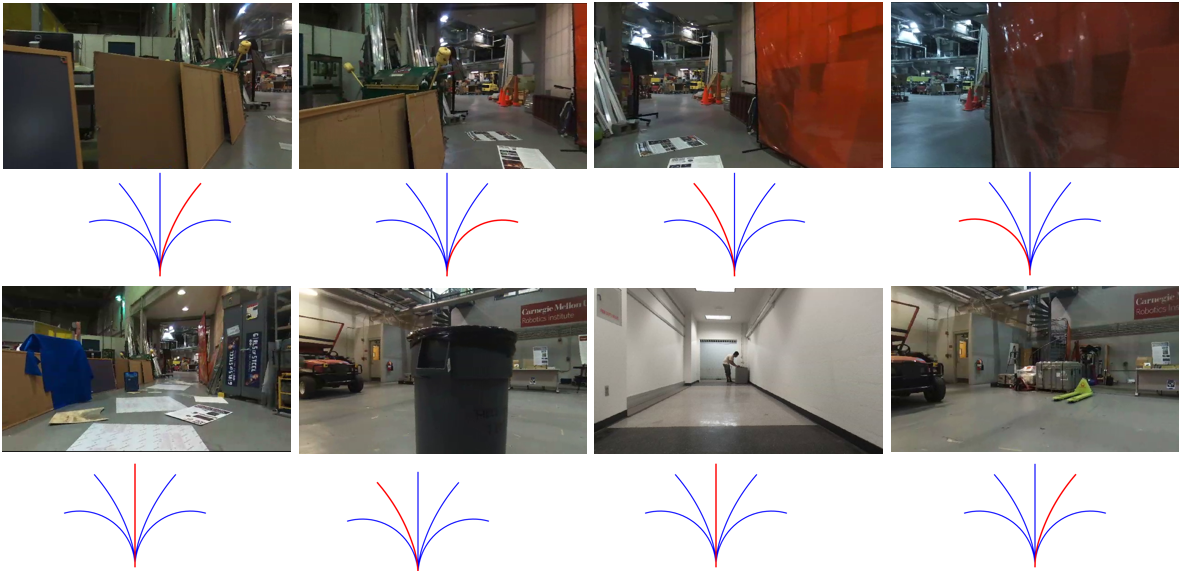}  
  \label{fig:real prediction}  
}
 \caption{ (a) Real flight scenes including curved corridor, front obstacles and corridor following (b) Eight prediction examples from quadrotor's view on the fly. For each image, the path image below it shows the predicted path. More results could be found in the supplementary video.
 }
 \label{fig:real flight}
\end{figure*}

We also test our algorithm on the real quadrobot flight. The platform is Parrot bebop quadrotor shown in Fig. \ref{fig:real scenes}. It can send the forward-facing camera image via WiFi to our host laptop, which predicts the trajectory and sends back velocity control command to bebop drone. The laptop is equipped with GeForce GTX 960M GPU for CNN prediction which takes about $0.143$s per frame. There is also image transmission delay about $0.2$s. We still use the trained model from \textit{NYUv2} dataset. We need to note that there are some other important factors affecting the actual flight performance such as state estimation and control. Due to instable and sometimes jumping state estimation from bebop drone, we cannot use the position feedback in the trajectory tracking which deteriorates the whole performance. We test our vehicle in three indoor environments: straight corridor, turning, front obstacles shown in Fig. \ref{fig:real scenes}.  Some prediction images on the fly are shown in Fig. \ref{fig:real prediction}. More results could be found in the supplementary video. We can see the robots can predict a reasonable trajectory in these tasks most of the time.

\section{Conclusion}
\label{sec:Conclusion}
In this paper, we propose a CNN based navigation system applicable to various environments. We start by training our system on public \textit{NYUv2} dataset instead of human demonstration datasets limited to certain scenarios in existing works. By designing 3D cost functions, we are able to automatically choose the best trajectory from each RGB-D image. Previous methods usually predict steering commands for robots which is difficult to evaluate whether it hits obstacles or not. Instead, we choose to predict a 3D trajectory which contains more meaningful 3D information.

Our perception system is also different from existing methods. We first instruct CNN to predict depth and surface normals which give the robot a sense of 3D obstacle position and scene layout and thus it can predict the path more precisely. Results on the \textit{NYUv2} dataset and other public datasets show that this intermediate perception performs much better and robustly than direct path prediction from raw RGB image.

We also apply our CNN model for quadrotor simulation and actual flight in different environments without retraining. The robot is able to avoid various obstacles in real time. Some future works include integrating it with high level motion planning, accurate state estimation and also adding more trajectories into it for example U-turn primitives. Integrating multiple frame's prediction should also improve the performance.

\bibliographystyle{unsrt}    % reference order according to appearance
\bibliography{ref}

\end{document}